\documentclass{article}
\usepackage{ijcai25}   %

\usepackage{times}
\usepackage{soul}
\usepackage{url}
\usepackage[hidelinks]{hyperref}
\usepackage[utf8]{inputenc}
\usepackage[small]{caption}
\usepackage{graphicx}
\usepackage{amsmath}
\usepackage{amsthm}
\usepackage{booktabs}
\usepackage{algorithm}
\usepackage{algorithmic}
\usepackage[switch]{lineno}

\usepackage{multirow}
\usepackage{tabularx} %
\graphicspath{{imgs/}}

\usepackage{amsfonts}       
\usepackage{nicefrac}       
\usepackage{microtype}      
\usepackage{xcolor}         
\theoremstyle{definition}

\usepackage{amssymb}
\usepackage{adjustbox}
\usepackage{bm}
\usepackage{multirow}
\usepackage{makecell}
\usepackage{wrapfig}

\usepackage{threeparttable}

\title{EFormer: An Effective Edge-based Transformer for Vehicle Routing Problems}

\author{
Dian Meng$^{1,4}$\and
Zhiguang Cao$^2$\and
Yaoxin Wu $^3$ \and
Yaqing Hou$^{1,4}$\footnotemark \and
Hongwei Ge$^{1,4}$\And
Qiang Zhang$^{1,4}$\\
\affiliations
$^1$School of Computer Science and Technology, Dalian University of Technology (DUT)\\
$^2$School of Computing and Information Systems, Singapore Management University\\
$^3$Department of Industrial Engineering and Innovation Sciences, Eindhoven University of Technology \\
$^4$Key Laboratory of Social Computing and Cognitive Intelligence (DUT), Ministry of Education, China\\
\emails
mengdian@mail.dlut.edu.cn,
zhiguangcao@outlook.com,
wyxacc@hotmail.com,
\{houyq, gehw, zhangq\}@dlut.edu.cn
}

\begin{document}

\maketitle

\begin{abstract}  
  Recent neural heuristics for the Vehicle Routing Problem (VRP) primarily rely on node coordinates as input, which may be less effective in practical scenarios where real cost metrics—such as edge-based distances—are more relevant. To address this limitation, we introduce EFormer, an Edge-based Transformer model that uses edge as the sole input for VRPs. Our approach employs a precoder module with a mixed-score attention mechanism to convert edge information into temporary node embeddings. We also present a parallel encoding strategy characterized by a graph encoder and a node encoder, each responsible for processing graph and node embeddings in distinct feature spaces, respectively. This design yields a more comprehensive representation of the global relationships among edges. In the decoding phase, parallel context embedding and multi-query integration are used to compute separate attention mechanisms over the two encoded embeddings, facilitating efficient path construction. We train EFormer using reinforcement learning in an autoregressive manner. Extensive experiments on the Traveling Salesman Problem (TSP) and Capacitated Vehicle Routing Problem (CVRP) reveal that EFormer outperforms established baselines on synthetic datasets, including large-scale and diverse distributions. Moreover, EFormer demonstrates strong generalization on real-world instances from TSPLib and CVRPLib. These findings confirm the effectiveness of EFormer’s core design in solving VRPs.  
\end{abstract}

\renewcommand{\thefootnote}{*}
\footnotetext{Yaqing Hou is the Corresponding author.}
 
\section{Introduction}
\renewcommand{\thefootnote}{1}

The vehicle routing problem (VRP), a fundamental NP-hard combinatorial optimization problem (COP), appears in numerous real-world contexts, including logistics~\cite{konstantakopoulos2022vehicle}, navigation systems~\cite{elgarej2021optimized}, and circuit design~\cite{brophy2014principles}. Despite extensive research across various fields, VRPs remain notably challenging due to their inherent computational complexity~\cite{ausiello2012complexity}. Approaches to solving VRPs can be divided into exact algorithms and heuristic algorithms~\cite{helsgaun2017extension}. Exact algorithms, although theoretically robust, often face scalability issues when applied to large instances because of their high computational demands. In contrast, heuristic algorithms are generally more practical yet depend heavily on manually crafted rules and domain-specific knowledge, which restricts their applicability and generalizability.

Recently, there has been a surge in neural heuristics that leverage deep (reinforcement) learning to solve VRPs. These methods learn problem-solving strategies end-to-end from data, offering a novel and efficient perspective on VRPs~\cite{kool2019AM}. Compared to traditional heuristics, neural heuristics typically provide higher solution efficiency and stronger generalization capabilities~\cite{zhang2025adversarial}.

Nevertheless, most existing neural heuristics rely heavily on node coordinates (or their embeddings) as a crucial input to the model, with many methods generating problem instances from these coordinates to train neural networks. The core assumption is that the relationship between coordinates and distance can be readily learned by the network when searching for the shortest path, particularly in classical Euclidean spaces. In these approaches, each iteration selects a node from the problem instance, and the encoded information of the remaining nodes is then used to incrementally construct the solution by inferring the corresponding Euclidean distance. However, this prevalent focus on node coordinates often lacks the robustness and generalizability required in practical applications. When distances cannot be easily inferred from coordinates alone, such methods tend to struggle, reducing their effectiveness in solving VRPs—especially in scenarios where the input space deviates from idealized conditions.

To address this issue, we propose a neural heuristic featuring an Edge-based Transformer (EFormer) model that exclusively relies on edge information as the original input. Specifically, we introduce a precoder module that employs a multi-head mixed-score attention mechanism to convert edge information into temporary node embeddings. We then design a parallel encoding strategy comprising two encoders: a graph encoder and a node encoder. The graph encoder employs a residual gated graph convolution network (GCN) to process sparse graph embeddings, while the node encoder leverages attention mechanisms to independently process temporary node embeddings. By encoding these two embedding types in separate feature spaces, the model achieves a more comprehensive representation of global edge relationships. Finally, parallel context embedding and multiple-query integration are used to decode the two types of encoded features, facilitating the effective construction of a complete path. The EFormer is trained using reinforcement learning in an autoregressive manner and achieves favorable performance on the TSP and CVRP. 
Accordingly, our contributions can be summarized as follows:
\begin{itemize}
    \item \textbf{Novel Edge-based Transformer (EFormer).} We propose a new and practical edge-based Transformer model designed to leverage the edge information for solving VRPs. The introduction of a multi-head mixed-score attention mechanism enables the extraction of temporary node features directly from edge weights.

    \item \textbf{Parallel Encoding Strategy.} We develop a parallel encoding strategy that utilizes residual gated graph convolution networks to encode graph embeddings and attention mechanisms to process node embeddings. By encoding graph and node embeddings in separate feature spaces, the model generates more comprehensive global edge relationship embeddings. A multiple-query integration method is then used to decode the embeddings, facilitating complete path construction.

    \item \textbf{Favorable Performance and Versatility.} Our purely edge-based method demonstrates strong performance on TSP and CVRP, surpassing other edge-based neural methods. Its robust generalization is evident in its performance on real-world instances from TSPLib and CVRPLib across diverse scales and distributions. The edge-based nature of EFormer also allows us to apply it to Asymmetric TSP (Appendix~\ref{appendix D}). Furthermore, we demonstrate its versatility by adapting it to solve VRPs using only node information as input (Appendix~\ref{appendix C}).
    
\end{itemize}

\section{Related Work}

\begin{figure*}[!t]
    \centering
    \includegraphics[width=0.90 \linewidth]{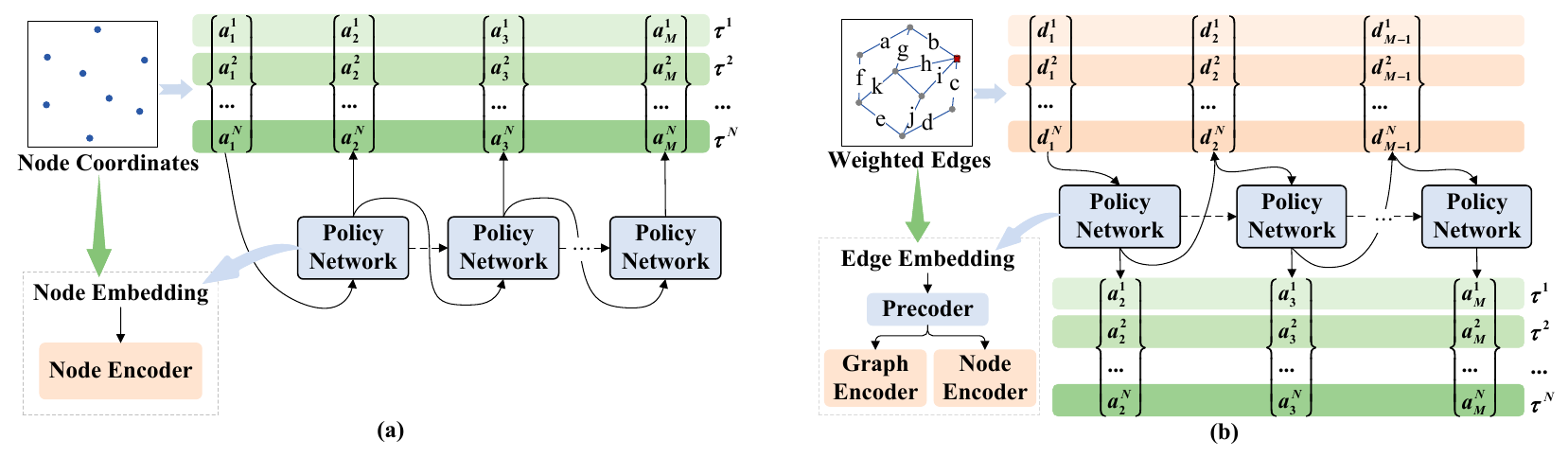}
    \caption{Comparison of node-based and edge-based policy networks, where multiple trajectory $\{\tau^1, \tau^2, \ldots, \tau^N\}$ is pursued in both for inferring solutions. (a) Overall scheme of the node-based methods such as POMO. (b) Overall scheme of the EFormer, which takes edge weights as the input.}
    \label{fig1:EFormer-XE}
\end{figure*}

\subsection{Node-based Neural Heuristics for VRP}

\paragraph{Graph neural network-based methods.} Graph neural networks (GNNs)~\cite{scarselli2008graph} provide a flexible framework for learning message-passing strategies among nodes, making them applicable to graphs of arbitrary size. In the context of routing problems, GNNs typically predict edge probabilities or scores, which are then leveraged by search algorithms (e.g., beam search, tree search, or guided local search) to produce approximate solutions~\cite{khalil2017learning,li2018combinatorial,nowak2017note,joshi2019efficient,fu2021generalize,xin2021neurolkh,hudson2021graph,kool2022deep,min2024unsupervised}.
\cite{khalil2017learning} proposed one of the earliest unified frameworks that combine reinforcement learning with graph embeddings to solve various combinatorial optimization (CO) problems on graphs. In another line of work, \cite{li2018combinatorial} incorporated advanced Graph Convolutional Networks (GCNs)~\cite{kipf2016semi} alongside tree search to explore the solution space for CO problems. \cite{nowak2017note} applied supervised learning to train a GNN and employed beam search to obtain feasible solutions. \cite{joshi2019efficient} designed a GCN model to predict a “heatmap” of edge probabilities in TSP instances, which guides beam search to produce feasible solutions.

\paragraph{Transformer-based constructive methods.} Among various neural construction heuristics, the Transformer~\cite{vaswani2017attention} represents a major breakthrough and has progressively become the leading approach for solving VRPs. These methods typically construct solutions incrementally by selecting one node at a time from the problem instance. Representative methods in this category include~\cite{nazari2018reinforcement,kim2021learning,kool2019AM,kwon2020pomo,jin2023pointerformer,drakulic2023bq,luo2023neural,huang2025rethink,lin2024cross}. Specifically, \cite{kool2019AM} is the first to leverage the Transformer architecture in a method called Attention Model (AM), thereby introducing a more powerful neural heuristic for VRPs. One notable variant of AM is POMO~\cite{kwon2020pomo}, which applies multiple optimal policies to significantly enhance AM’s learning and inference capabilities. More recently, \cite{luo2023neural} proposed the Light Encoder and Heavy Decoder (LEHD) model, trained via supervised learning on 100-node instances. LEHD not only achieves higher-quality solutions but also demonstrates strong generalization capabilities.

\paragraph{Transformer-based improvement methods.} Neural improvement heuristics iteratively refine an initial feasible solution until a stopping criterion (e.g., convergence) is reached. Drawing inspiration from classical local search algorithms, they can optimize sub-problems or apply improvement operators (e.g., k-opt) to enhance solution quality. Representative improvement-based approaches include ~\cite{lu2019learning,barrett2020exploratory,wu2021learning,ma2021learning,li2021learning,wang2021bi,kim2023learning,cheng2023select}.

\subsection{Edge-based Neural Heuristics for VRP}
Most neural heuristics, typically based on GNNs or Transformers, capture the structure of routing problems by treating the coordinates of problem instances as node features (Figure~\ref{fig1:EFormer-XE}, (a)). However, encoding edge features rather than relying solely on node features more closely aligns with practical applications. Early works incorporating edge features include variants of Graph Attention Networks (GAT)~\cite{velivckovic2017graph}. For example, \cite{chen2021edge} introduced Edge-featured Graph Attention Networks (EGAT), which consider edge features during message-passing, while \cite{shi2020masked} integrated edge features into attention-based GNNs using “Graph Transformers” for semi-supervised classification. In addition, \cite{jin2023edgeformers} proposed EdgeFormers, an architecture that processes text edge networks to enhance GNNs for better utilization of edge (text) features.

In the context of routing problems, several edge-based approaches have been explored, such as MatNet~\cite{kwon2021matrix} and GREAT~\cite{lischka2024great}. Specifically, \cite{kwon2021matrix} proposed a Matrix Encoding Network (MatNet) that accepts an encoded distance matrix to solve complex asymmetric traveling salesman (ATSP) and flexible flow shop (FFSP) problems. Meanwhile, \cite{lischka2024great} introduced the Graph Edge Attention Network (GREAT), an edge-based neural model related to GNNs, which uses highly sparse graphs to achieve high-quality solutions. Following this line of research, our work also focuses on edge-based methods and proposes a neural network model architecture that improves the solution quality for solving VRPs (Figure~\ref{fig1:EFormer-XE}, (b)).


\section{Methodology: EFormer}
We introduce the EFormer, a model for solving VRPs that comprises four main components: a \emph{precoder}, a \emph{graph encoder}, a \emph{node encoder}, and a \emph{decoder}. Given a set of edge information, EFormer first applies the precoder to generate temporary node embeddings. To mitigate the heavy computation arising from dense graphs, we adopt a k-nearest neighbor (k-nn) heuristic to sparsify them. It then employs parallel graph and node encoders to process graph and node embeddings in distinct feature spaces, respectively. Finally, the decoder constructs a path by integrating parallel context embeddings and a multi-query mechanism. Figure~\ref{fig2:overall} shows the overall framework of EFormer, while a more detailed version is provided in Appendix~\ref{appendix A.1}.

\begin{figure*}[!t]
    \centering
    \includegraphics[width=1.0 \linewidth]{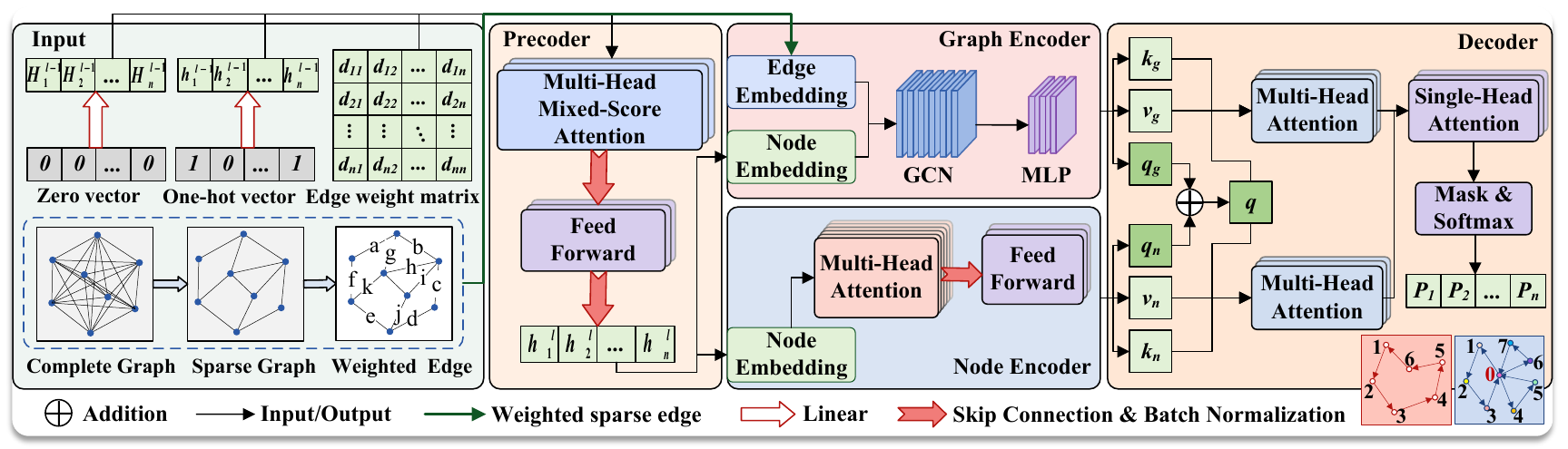}
    \caption{Overview of the EFormer framework. From the input of sparse weighted edges, the \emph{precoder} performs preprocessing, the \emph{graph encoder} and \emph{node encoder} operate in parallel, and finally the \emph{decoder} constructs the complete path.}
    \label{fig2:overall}
\end{figure*}

\subsection{Precoder}
The precoder processes externally provided edge information to generate node embeddings, effectively serving as a pre-encoding step in EFormer. It comprises a multi-head mixed-score attention layer and a feed-forward (FF) layer, as shown in Figure~\ref{fig2:overall}. We introduce its key steps below.

\paragraph{The input's initial representations.} Along with the edge weights, we also incorporate a zero vector, a one-hot vector, and the edge weight matrix. This approach is crucial because the one-hot vector is randomly selected from a predefined matrix pool, ensuring that each iteration generates unique embeddings. Consequently, we can supply the same problem instances along with the zero embedding multiple times, enabling flexible instance augmentation.

We generally follow the graph attention networks (GATs) framework~\cite{velivckovic2017graph}, which is described in detail in Appendix~\ref{appendix A.2}. Nevertheless, unlike the classic GATs, the attention score for a pair of nodes $(i, j)$ in our method depends not only on $\hat{h}_{i}$ and $\hat{h}_{j}$, but also on the edge weight $e(i,j)$. 
The precoder’s update function is defined as: 
\begin{align} \hat{h}_{i}' = \mathcal{F}_R\Big(\hat{h}_{i}, \hat{h}_{j}, {e(i,j)\mid i \in R, j \in M}\Big), \end{align}
where the learnable update function $\mathcal{F}$ leverages multi-head attention (MHA). Its aggregation process uses attention scores for each node pair $(i, j)$, which are determined by $\hat{h}_{i}$ and $\hat{h}_{j}$. Here, $R$ denotes the set of adjacent nodes of $i$, and $M$ denotes the set of adjacent nodes of $j$. 
  
\paragraph{Mixed-score attention.} 
For inputs containing only edge information, it is crucial to incorporate edge weights $e(i, j) \equiv D_{ij}$ in the attention mechanism. Drawing inspiration from \cite{kwon2021matrix}, we adopt a “Multi-Head Mixed-Score Attention” block to process the edge weight matrix. This block closely follows the MHA module in Transformer, except that the scaled dot-product attention in each head is replaced with mixed-score attention. Specifically, the block integrates the externally provided edge distance matrix $D_{ij}$ with the internally generated attention scores. A small Multilayer Perceptron (MLP) with two inputs and one output determines the optimal way to combine these scores. The resultant mixed scores then pass through a “softmax” stage, thereby retaining matrix-based representation crucial to attention mechanisms.
 
By performing mixed-score attention on the given edge information, it produces an encoded relationship matrix $h_{i}^{(P)}$, which acts as the node embeddings for both the graph and node encoders. This process is given by:
\begin{align}
\hat{h_i}^{(P)} &= h_r + \operatorname{mixed-scoreMHA}(h_r,h_c,D_{ij}), \\
h_i^{(P)} &= \hat{h_i}^{(P)} + \operatorname{FF}(\hat{h_i}^{(P)}),
\end{align} 
where $D_{ij}$ is edge weight matrix, $h_r$ is zero-vector embedding, and $h_c$ is one-hot vector embedding.

\paragraph{Instance augmentation.}
Since we initialize each run with a random sequence of one-hot vectors, the weight matrix $D_{ij}$ is encoded differently each time. As a result, the precoder can yield diverse representations of the same problem instance and generate distinct solutions. By providing a new one-hot vector sequence for each run, a multitude of different solutions can be readily obtained simply by repeatedly executing the model. Moreover, this random one-hot embedding strategy naturally aligns with the “instance augmentation” technique proposed in POMO~\cite{kwon2020pomo}. However, whereas POMO provides ×8 instance augmentation, EFormer can achieve ×N augmentation. The trade-off is that while this approach yields higher-quality solutions, it also increases runtime. 

\subsection{Graph Encoder} 
The graph encoder is a pivotal component of our parallel strategy, comprising residual gated GCN layers and MLP layers. It processes a sparse graph formed by the edge embeddings and node embeddings produced by the precoder.
\paragraph{Graph sparsification.} To mitigate the computational burden arising from dense graphs, we employ a k-nn heuristic, in which each node is connected to its $k$ nearest neighbors. The choice of $k$ and its impact on training efficiency are evaluated in the experimental section, where we demonstrate the importance of sparsity for improving training efficiency.
\paragraph{Input layer.} 
The precoder’s output $h_i^{(P)}$, serves as the graph encoder’s node embedding. We project $h_i^{(P)}$ into $h$-dimensional feature as $x^{l=0}_i = A_1 h_i^{(P)}$ with $A_1 \in R^{h}$. 

We define an edge adjacency matrix function $\delta_{ij}^{\operatorname{KNN}}$, which has a value of 1 if nodes $i$ and $j$ are $k$-nearest neighbors, a value of 2 if they are self-connected, and a value of 0 otherwise. The edge adjacency matrix $\delta_{ij}^{\operatorname{KNN}}$ and the edge weight matrix $D_{ij}$ are embedded as $\frac{h}{2}$-dimensional feature vectors. Then we concatenate the two together to get the edge input feature as $
e^{l=0}_{ij} = A_2 D_{ij} + b_3||A_3 \cdot \delta_{ij}^{\operatorname{KNN}}$, where $A_2 \in R^\frac{h}{2}$, $A_3 \in R^\frac{h}{2}$, $b_3$ is the bias, and $\cdot||\cdot$ is the concatenation operator. 

\paragraph{Residual gated Graph Convolution layer.}
Let $x^l_i$ and $e^l_{ij}$ represent the node and edge embedding associated with the node $i$ and edge $(i,j)$ at layer $l$, respectively. In our method, we employ the GCN architecture introduced in~\cite{bresson2017residual}, to produce the node embedding $x_{i}^{l+1}$ and edge embedding $e_{ij}^{l+1}$ at the next layer as follows:
\begin{align}
x_i^{l+1} &= x_i^{l} + \operatorname{ReLu}(\operatorname{BN}(W_1^{l} x_i^{l} + \eta_{ij}^{l} \odot W_2^{l} x_j^{l} )),\\
e_{ij}^{l+1} &= e_{ij}^{l} + \operatorname{ReLu}(\operatorname{BN}(W_3^{l} e_{ij}^{l} + W_4^{l} x_i^{l} + W_5^{l} x_j^{l})), \\
\eta_{ij}^{l} &= \sum_{j\sim i} \frac{\sigma(e_{ij}^{l})}{\sum_{j'\sim i}\sigma(e_{ij'}^{l})+\xi},
\end{align} 
where $W^l_* \in R^{h}$, $\sigma$ is the sigmoid function, $\xi$ is a small value, ReLU is the rectified linear unit, and BN stands for batch normalization~\cite{ioffe2015batch}. Meanwhile, GCN exploits the additive edge feature representation and dense attention map $\eta_{ij}$, to make the diffusion process anisotropic on the graph. 

\paragraph{MLP layer.}
In the final layer, MLP modules process both the node and edge embeddings derived from the GCN layers. Here, we specifically focus on the node embedding $x_i^l$, which inherently incorporates the relevant edge and node information. The MLP outputs $h^G_{ij}$ in $[0,1]^{2}$ defined as $h^G_{ij} = \operatorname{MLP}(x_i^{l})$, where the number of MLP layers is denoted by $l_{mlp}$.

\subsection{Node Encoder}
Node encoder processes the node embeddings generated by precoder, using attention mechanisms to embed them into the distinct feature space and produce encoded representations. 

\paragraph{Input layer.}
Unlike the traditional approach~\cite{kool2019AM} that relies on node coordinates, the node encoder uses the temporary node embedding $h_{i}^{(P)}$ derived from the precoder as the input to its attention layer. Let $H^{(0)}$ denote the input to the first layer, then it will be given as $H^{(0)}= h_{i}^{(P)}$.

\paragraph{Attention layer.}
Node encoder comprises $n$ attention layers. Each attention layer consists of two sublayers: a MHA sublayer and a FF sublayer~\cite{vaswani2017attention}. Each sublayer incorporates a residual connection~\cite{he2016deep} and batch normalization (BN).
We denote the embedding obtained from each layer as $h_i^{(l)}$, and let \( H^{(l-1)} = \left( h_1^{(l-1)}, h_2^{(l-1)}, \ldots, h_n^{(l-1)} \right) \) 
be the input to the $l$-th attention layer when $l = \{1,..., L\}$. The output of the attention layer for the embedding of the $i$-th node is calculated as follows:
\begin{align}
\hat{h}_i^{(l)} &= \operatorname{BN}\left(h_i^{(l-1)} + \operatorname{MHA}\left(h_i^{(l-1)}, H^{(l-1)}\right)\right),\\
h_i^{(l)} &= \operatorname{BN}\left(\hat{h}_i^{(l)} + \operatorname{FF}\left(\hat{h}_i^{(l)}\right)\right),
\end{align}
where the FF sublayer includes one hidden sublayer and ReLU activation. The above process and the final output can be summarized as follows:
\begin{align}
H^{(l)} &= \operatorname{AttentionLayer}(H^{(l-1)}), 
\quad
h^N_{L} = H^{(L)}.
\end{align}

\emph{\textbf{Note.}} we employ two parallel encoding components—the graph encoder and node encoder—both of which take node embeddings as (one of the) inputs. One might wonder why the node encoder is needed if the graph encoder can already handle node embeddings. Although using multiple GCN layers can yield powerful encoding representations, it often increases model complexity. By contrast, the node encoder uses fewer layers and leverages attention mechanisms, allowing the model to remain lightweight while still delivering strong performance. Moreover, the attention mechanisms have been shown to be highly effective for encoding node embeddings. Empirically, this parallel architecture outperforms approaches that rely solely on GCNs or a single attention-based encoder.

\subsection{Decoder}
During the decoding phase, the edge embeddings and node embeddings produced by the graph encoder and node encoder, respectively, are processed in parallel. We use superscripts to differentiate the sources of information: $G$ denotes edge embeddings, while $N$ denotes node embeddings. Initially, two separate sets of keys and values are extracted from the edge and node embeddings. In parallel, the current contextual embedding is derived from the current state in combination with these two sets of embeddings.

The embeddings of all the starting nodes $h_{first}^{G}$, $h_{first}^{N}$ (i.e., the nodes selected in the first step) and the embeddings of the target nodes $h_{last}^{G}$, $h_{last}^{N}$ (i.e., the currently selected nodes) are concatenated to form two sets of temporary queries (i.e., $q^{G}$, $q^{N}$). Afterwards, the two sets of context embeddings are merged to create the final query as follows:
\begin{align}
q^{G} &= W_1^{G} h_{first}^{G} + W_2^{G} h_{last}^{G}, \\
q^{N} &= W_1^{N} h_{first}^{N} + W_2^{N} h_{last}^{N}, \\
q &= q^{G} + q^{N},
\end{align}
where the subscript ${first}$ denotes the fixed start node, and the subscript $last$ denotes the current target node. $W_1^{G}, W_2^{G}, W_1^{N}$ and $W_2^{N}$ are learnable matrices used to recast the start node embeddings $h_{first}^{G}$, $h_{first}^{N}$ and the current target node embeddings $h_{last}^{G}$, $h_{last}^{N}$, respectively. 

Next, we apply the MHA mechanism to each set of context embeddings separately, producing two outputs, $A^G$ and $A^N$: 
\begin{align}
A^{G} &= \operatorname{MHA}(q, k^{G}, v^{G}),\\
A^{N} &= \operatorname{MHA}(q, k^{N}, v^{N}),
\end{align} 
We then apply two linear layers $W_3^G$ and $W_3^N$ to map $A^G$ and $A^N$, respectively. Subsequently, we compute a score via two sets of single-head attention layers, apply the tanh function to clip the score, and mask any visited nodes. The resulting score $u_j$ for node $j$ is given by:
\begin{align}
u_j = 
\begin{cases} 
C \cdot \tanh\left(\frac{(W_3^G A^G + W_3^N A^N)(k_j^G + k_j^N)}{\sqrt{d_k}}\right),\text{if } j \text{ unvisited} \\
- \infty, \text{otherwise}
\end{cases}
\end{align} 
where $W_3^G$ and $W_3^N$ are learnable matrices, $u_j$ is the score for node $j$, and $d_k$ is determined by the embedding dimension. We then apply the softmax function to calculate the probability $p_j$ of selecting node $j$. At each decoding step $j$, the next node is selected based on its probability $p_j$. Repeating this process $n$ times yields the complete solution \(\tau=\{\tau^1,\cdots,\tau^n\}^T\):  
\begin{align}
p_j &= \operatorname{softmax}(u_j).
\end{align}

\subsection{Training}
Since EFormer can be easily integrated into a variety of autoregressive solvers, we adopt the same reinforcement learning training method as POMO. 
We train the EFormer model using the REINFORCE algorithm~\cite{williams1992simple}. We sample a set of $n$ trajectories $\{\tau^1,\cdots,\tau^n\}$, calculate the reward $f(\tau^i)$ for each, and employ approximate gradient ascent to maximize the expected return $\mathcal{L}$.
The gradient of the total training loss $\mathcal{L}$ can be approximated as follows: 
\begin{align}
\nabla_\theta \mathcal{L}(\theta) \approx \frac{1}{n} \sum_{i=1}^{n}[(f(\tau^i)-b^i(s)) \nabla \log p_\theta(\tau^i|s)],
\end{align}
where $b^i(s)$ is commonly set as the average reward of those $m$ trajectories, serving as a shared baseline:
\begin{align}
b^i(s) &= b_{\text{shared}}(s) = \frac{1}{n} \sum_{i=1}^{n} f(\tau^i), \text{ for all } i \\
\text{where } p_\theta(\tau^i|s) &= \prod_{t=2}^{M} p_\theta(a_t^i|s,a_{1:t-1}^i).
\end{align}


\begin{table*}[!t]
    \centering
    \resizebox{0.83\linewidth}{!}{
    \begin{tabular}{lrrrrrrrrr}
        \toprule
         & \multicolumn{3}{c}{TSP20}&\multicolumn{3}{c}{TSP50}& \multicolumn{3}{c}{TSP100} \\   
        Method  & Len. & Gap($\%$) & Time($m$) & Len. & Gap($\%$) & Time($m$) & Len. & Gap($\%$) & Time($m$)    \\
        \midrule    
        Concorde& 3.831& 0.000&4.43 &5.692& 0.000 &23.53&7.763 &0.000 &66.45\\
        LKH3     & 3.831&0.000&2.78 &5.692&0.000 &17.21&7.763 &0.000 &49.56\\
        OR-Tools& 3.864& 0.864&1.16&5.851& 2.795 &10.75&8.057 &3.782 &39.05 \\
        \hline
        GCN-Greedy&3.948&3.078&0.32& 5.968&4.856&1.34 &8.537 &9.966& 4.09 \\
        GCN-BS &3.862 &0.825&0.81& 5.732 &0.712&4.33&8.170 &5.234 &4.31 \\
        GCN-BS*&3.831 &0.004& 21.41& 5.700&0.141&35.46&7.955 &2.477&63.05 \\
        MatNet(×1)&3.832 &0.044&0.11&5.709& 0.303&0.13&7.836& 0.940&0.52    \\
        MatNet(×8)&3.831 &0.002&0.22&5.694& 0.050&1.24&7.795 &0.410&5.28   \\
        MatNet(×128)&3.831 &0.000&5.71&5.692&0.013&16.47&7.776&0.170&60.11 \\
        GREAT(×1)\#& - & - &  - & - & - & - & 7.850& 1.210& 2.00     \\
        GREAT(×8)\#& - & -  & - & - & - & - & 7.820& 0.810& 18.00  \\
        \hline
       EFormer(×1)&3.831 &0.018&0.04 &5.699 &0.130&0.26 &7.788 &0.324& 1.22\\
       EFormer(×8)&3.831&0.000&0.15 &5.692 &0.011&1.34 &7.772&0.115 & 6.81  \\
       EFormer(×128) &3.831 &\textbf{0.000}&4.72 &5.692&\textbf{0.001}&25.81&7.767 & \textbf{0.045}&66.55 \\
        \midrule \midrule
      
      & \multicolumn{3}{c}{CVRP20} & \multicolumn{3}{c}{CVRP50}& \multicolumn{3}{c}{CVRP100} \\   
       Method  & Len. &Gap($\%$) & Time($m$) & Len. & Gap($\%$) & Time($m$) & Len. &Gap($\%$) & Time($m$) \\
        \midrule    
        LKH3    &6.117 & 0.000&2.15h &10.347& 0.000 &8.52h &15.647 &0.000 &13.46h\\
        HGS&6.112&-0.079&1.48h&10.347&-0.001 &4.67h &15.584&-0.401&6.54h\\
        OR-Tools&6.414 & 4.863& 2.37 &11.219& 8.430 &19.35 &17.172 &9.749 &2.61h \\\hline
        
        GCN-Greedy&6.471&5.794&0.27&11.130&7.567&2.05 &16.948&8.314&5.24 \\
        GCN-BS &6.284 &2.740&0.26&10.786&4.248&2.11  &16.487&5.371 & 5.45   \\
        GCN-BS*&6.192 &1.232& 20.78&10.636&2.796&38.36&16.243&3.811&78.78  \\
        
        MatNet(×1)&6.172 &0.907&0.11&10.787&4.253&0.21 &16.280&4.401&1.02\\
        MatNet(×8)&6.146 &0.469 &0.58 &10.635&2.787&1.23&16.117&3.356&4.70 \\
        MatNet(×128)&6.131&0.229 &9.93 &10.538&1.847&17.93&15.989&2.530&66.05 \\\hline
        
       EFormer(×1)&6.147 &0.490&0.04&10.457&1.067&0.25&15.844&1.259 &0.98 \\
      EFormer(×8)&6.123&0.098& 0.28&10.414&0.650&1.69 &15.776&0.830&6.86\\
    EFormer(×128)&6.116&\textbf{-0.017}&14.84&10.393&\textbf{0.447}&24.47&15.735 &\textbf{0.563} &85.65   \\ 
        \bottomrule
    \end{tabular}
    }
    \caption{Experimental results on TSP and CVRP with uniformly distributed instances. The results of methods with an asterisk (\#) are directly obtained from the original paper. BS: Beam search, BS*: Beam search and shortest tour heuristic}
    \label{tab1:vrp-edge}
\end{table*}

\begin{table*}[!t]
    \centering
    \footnotesize
    \resizebox{0.84\linewidth}{!}{
    \begin{tabular}{lrr|rr||lrr|rr}
        \toprule
   
     & \multicolumn{2}{c|}{TSP50} & \multicolumn{2}{c||}{CVRP50}  &
      & \multicolumn{2}{c|}{TSP50} & \multicolumn{2}{c}{CVRP50}
            \\   

        Method  & Len. & Gap(\%)  & Len. & Gap(\%) &
        Method  & Len. & Gap(\%)  & Len. & Gap(\%)  \\
        \midrule    
 
        OPT & 5.692& 0.000 &10.347&0.000 &
        OPT & 5.692& 0.000 &10.347&0.000 
        \\ 
        \midrule    
       
        K=10 (×1)&5.698&\textbf{0.117} &10.476&1.250 &
        w.o. precoder(×1) & 5.702&0.181 &10.642&2.854 
        \\
        K=20 (×1)&5.699&0.130 &10.474&\textbf{1.231} &
        w.o. node encoder(×1) &5.702&0.185 & 10.517 &1.640 

        \\
        K=30 (×1)&5.699&0.135 &10.484&1.321  &
        w.o. graph encoder(×1)&5.705&0.233 & 10.499 &1.466 
        
         \\
        K=40 (×1)&5.700&0.147 &10.486&1.346 &   
        w.o. gcn(×1)  &5.707&0.276 & 10.502 &1.501
         \\
        K=50 (×1)&5.699&0.133 &10.485&1.332 & 
        \textbf{EFormer(×1)}&5.699&\textbf{0.130} &10.474 &\textbf{1.231}
        \\
        \midrule    
        
        K=10 (×8) &5.692&0.012 &10.424&0.746 &  
        w.o. precoder(×8) & - & -           &- & -         
        \\
        
        K=20 (×8) &5.692&\textbf{0.011} &10.422&\textbf{0.725}& 
        w.o. node encoder(×8) &5.693&0.026 & 10.442 &0.918 
        \\
        K=30 (×8) &5.692&0.012 &10.423&0.734 & 
        w.o. graph encoder(×8)&5.693&0.020 & 10.431 &0.817 
        
        \\
        K=40 (×8) &5.693&0.020 &10.423&0.740 & 
        w.o. gcn(×8)  &5.693&0.021 & 10.428 &0.782 
        \\
        K=50 (×8) &5.692&0.017 &10.422&0.731& 
       \textbf{EFormer(×8)}&5.692&\textbf{0.011} &10.422 &\textbf{0.725}
        \\
        \bottomrule
    \end{tabular}   
    }
    \caption{Ablations of various K values and four key elements of EFormer on uniformly distributed instances.}
    \label{tab2:Ablation}
\end{table*}

\section{Experiment}
\renewcommand{\thefootnote}{1}

\begin{table*}[!t]
    \centering
    \resizebox{1.02\linewidth}{!}{
    \begin{tabular}{lcrcrcr||crcrcr}
        \toprule
         & \multicolumn{2}{c}{TSPLIB1-100} & \multicolumn{2}{c}{TSPLIB101-300} & \multicolumn{2}{c||}{TSPLIB301-500}
          & \multicolumn{2}{c}{CVRPLIB1-100} & 
          \multicolumn{2}{c}{CVRPLIB101-300} & \multicolumn{2}{c}{CVRPLIB301-500} \\
          Method & Len. & Gap(\%) & Len. & Gap(\%) & Len. & Gap(\%)  
          & Len. & Gap(\%) & Len. & Gap(\%) & Len. & Gap(\%) \\
        \midrule    
        OPT &19499.583& 0.000  &40129.375& 0.000 &43694.666 &0.000 
         &925.800 & 0.000 &33184.483& 0.000 &82903.857 &0.000\\ 
        
        GCN-Greedy&25371.528&30.242 &63125.529&55.716&81143.342&83.080
           &1233.255&34.732&44684.202&45.069&138044.859&63.597\\
        GCN-BS &23906.973&22.631 &58213.359&46.951 &76546.164&72.946
          &1167.058&27.909&42773.044&38.466&120532.679&49.664\\
        GCN-BS*&21195.267&10.824 &55853.973&39.104 &72865.950&64.648 
          &1087.529&18.369&41403.524&34.012&114962.535&42.366\\
        MatNet(×1)&20076.205&3.880&46876.768&17.868&63544.470&46.192
          &1098.828 &21.017 & 38946.522 &20.351 &103278.318 &24.309\\
        MatNet(×8)&19761.313&1.786&45638.627&14.479&62427.744&42.206
          &1072.623&14.901 & 38218.037 &17.657 & 98912.043&19.460
        \\
        \midrule    
        
        EFormer(×1)&19741.575&1.698 &45396.807&13.599&61415.142&36.562 
        &1107.504&18.766&38477.709&12.485&98010.813&21.023\\
        EFormer(×8)&19662.845&\textbf{0.837} &43360.592&\textbf{8.805}&56111.793&\textbf{26.924} 
        &1019.698&\textbf{9.885} &36968.746&\textbf{9.784}&94612.575&\textbf{15.325}\\
  \midrule 
     \end{tabular}
    }
    \caption{Experimental results on TSPLib and CVRPLib.}
    \label{tab5:TSPLIB-CVRPLIB}
\end{table*}

We empirically evaluate our proposed EFormer model on TSP and CVRP of various sizes and distributions, comparing it against both learning-based and classical solvers. Our code is publicly available\footnote{\url{https://github.com/Regina921/EFormer}}. 
\paragraph{Basic Settings.}
We compare EFormer with: \textbf{(1) Classical solvers:} Concorde~\cite{cook2011traveling}, LKH3~\cite{helsgaun2017extension}, HGS~\cite{vidal2022hybrid}, and OR-Tools~\cite{ORTOOLS}; \textbf{(2) Heatmap-based method:} GCN-BS~\cite{joshi2019efficient}; \textbf{(3) Edge-based neural heuristics:} MatNet~\cite{kwon2021matrix} and GREAT~\cite{lischka2024great}. We follow the standard data generation procedures from prior work~\cite{kool2019AM} to create training and testing datasets for TSP and CVRP, where distances between nodes are calculated and provided as inputs accordingly. Each training epoch samples 100,000 random instances, while a separate set of 10,000 uniformly generated instances is used for testing. Optimal solutions for TSP are obtained via the Concorde solver, and those for CVRP using LKH3. We adopt the POMO inference algorithm~\cite{kwon2020pomo} and report both the optimality gap and inference time. For EFormer specifically, we present results for greedy inference (×1) and instance augmentation (×8 and ×128). Details of the experimental setup and additional baseline information can be found in Appendix~\ref{appendix B.1}.

Table~\ref{tab1:vrp-edge} presents our main results on uniformly distributed TSP and CVRP instances. For TSP, our proposed EFormer achieves excellent greedy inference (x1) performance across various instance sizes, all within a relatively reasonable inference time. Additionally, we perform inferences with instance augmentation (×8, ×128), which significantly outperform other neural heuristics. Notably, our x8 augmentation even surpasses MatNet’s x128 augmentation. For TSP100, the optimality gap can be as low as 0.0453\% when using x128 augmentation. EFormer thus clearly outperforms other edge-based methods, whether using greedy inference or instance augmentation.

For CVRP, we extend and refine the two learning-based baselines from their original formulations while retaining their model structure and parameters to ensure effective CVRP solutions. Regardless of whether greedy inference (×1) or instance augmentation (×8 or ×128) is employed, our EFormer consistently outperforms other edge-based methods. With ×128 augmentation, EFormer achieves an optimality gap of 0.5633\% on CVRP100. Thus, EFormer demonstrates superior performance over competing approaches under both greedy inference and augmented settings. Given a larger time budget, EFormer can leverage additional instance augmentation to further enhance performance across all instances.

\subsection{Ablation Study}
We conduct two sets of ablation studies to clarify essential design choices in our method. Specifically, we focus on: (1) the selection of the hyperparameter $K=20$ for the k-nn graph sparsification, and (2) the necessity of using three encoders.

\paragraph{$K$ value selection.}
In the classic knn-based sparsification approach, the hyperparameter $K$  determines how many edges are retained for each node ($K\times N$, where $N$ is the TSP size). We compare various $K$ values (10, 20, 30, 40, and 50) on TSP50 and CVRP50; the results are shown in Table~\ref{tab2:Ablation}. For TSP50, the optimality gaps for $K$ = 10, 20, and 30 are similar, with $K$ = 20 performing slightly better. For CVRP50, $K$ = 20 produces the highest solution quality. Overall, $K$ = 20 appears to capture nearly all the best solutions for TSP50 and CVRP50, so we set $K$ = 20 in all subsequent experiments.

\paragraph{The necessity of three encoders.}
Table~\ref{tab2:Ablation} shows the ablation results for EFormer and four variants. The first variant removes the precoder, retaining only the node encoder and graph encoder (denoted by w.o. precoder in the table). Without precoder, the model cannot re-encode the same problem instance multiple times; hence, no additional instance augmentation is performed. The second variant removes the node encoder, keeping only the precoder and graph encoder (denoted by w.o. node encoder). The third variant removes the graph encoder, retaining the precoder and node encoder (denoted by w.o. graph encoder). The fourth variant removes the GCN module from the graph encoder but retains the precoder, node encoder, and MLP module of graph encoder (denoted by w.o. gcn). Comparing the third and fourth variants highlights the effectiveness of our parallel dual-encoder structure. As shown in Table~\ref{tab2:Ablation}, EFormer outperforms all variants, indicating that each component contributes positively to the model.


\subsection{Generalization}
We assess the generality of our proposed EFormer from three perspectives: 1) real-world TSPLIB and CVRPLIB benchmarks, 2) larger-scale instances, and 3) different distributions. 

\paragraph{Generalization to TSPLIB and CVRPLIB.} 
Table~\ref{tab5:TSPLIB-CVRPLIB} summarizes the results on real-world TSPLIB~\cite{reinelt1991tsplib} and CVRPLIB~\cite{uchoa2017new} instances of various sizes and distributions. EFormer performs best on instances with up to 100 nodes and ranks second for 101–300 nodes, demonstrating excellent generalization on both TSPLIB and CVRPLIB. It also outperforms GCN-BS and MatNet on all instances.

\paragraph{Generalization to larger-scale instances.} 
Appendix~\ref{appendix B.2} shows the performance of EFormer on TSP and CVRP instances with up to 500 nodes. Despite device limitations, EFormer significantly surpasses other edge-based methods, highlighting its robust generalization even when only edge information is available.

\paragraph{Generalization across different distributions.} We further evaluate EFormer on TSP and CVRP instances from explosion, grid, and implosion distributions. Appendix~\ref{appendix B.3} presents the results, indicating that EFormer consistently outperforms GCN-BS and MatNet across all three distributions. Moreover, it maintains strong performance not only on the uniform distribution but also under varying distribution scenarios, underscoring its solid generalization capabilities.

\subsection{Node-based EFormer}
Our EFormer architecture is highly flexible, enabling it to address VRPs using node coordinates as the only inputs. To compare the performance of EFormer-based solvers with other established methods, we introduce a variant called EFormer-node, which is tested on the traditional node-coordinate setting. Experimental results indicate that EFormer-node delivers competitive performance compared to other established neural heuristics. For further details on the model architecture and experimental findings, please refer to Appendix~\ref{appendix C}. Additionally, we also solve ATSP based on our EFormer framework, and the detailed experimental results are presented in Appendix~\ref{appendix D}.

\section{Conclusion}
In this paper, we introduce a novel Edge-based Transformer (EFormer) model designed to solve VRPs in an autoregressive way that utilizes edge information as input. Our integrated architecture employs three encoders that work in concert to efficiently capture and process edge information. By adopting a parallel encoding approach, we encode different types of information in separate feature spaces, thereby enhancing the global strategy. Extensive experiments on both uniformly generated synthetic instances and real-world benchmarks demonstrate EFormer’s strong performance. Compared to node-based approaches, edge-based methods exhibit greater flexibility and applicability to real-world scenarios. Looking ahead, we plan to investigate lightweight architectural designs for EFormer to improve its scalability across a broader range of problem sizes. Another promising direction is to develop a unified learning-based framework that can operate effectively on both edge and node, where both are available as inputs.

\section*{Acknowledgments}
This work was supported in part by the National Natural Science Foundation of China under Grant 62372081, the Young Elite Scientists Sponsorship Program by CAST under Grant 2022QNRC001, the Liaoning Provincial Natural Science Foundation Program under Grant 2024010785-JH3/107, the Dalian Science and Technology Innovation Fund under Grant 2024JJ12GX020, the Dalian Major Projects of Basic Research under Grant 2023JJ11CG002 and the 111 Project under Grant D23006.
This research is supported by the National Research Foundation, Singapore under its AI Singapore Programme (AISG Award No: AISG3-RP-2022-031).


\bibliographystyle{named}
\bibliography{ijcai25}

\newpage
\appendix

\section{EFormer Model}
\label{appendix A}

\subsection{EFormer Detailed Architecture}
\label{appendix A.1}

\begin{figure*}[!t]
    \centering
    \includegraphics[width=0.85 \linewidth]{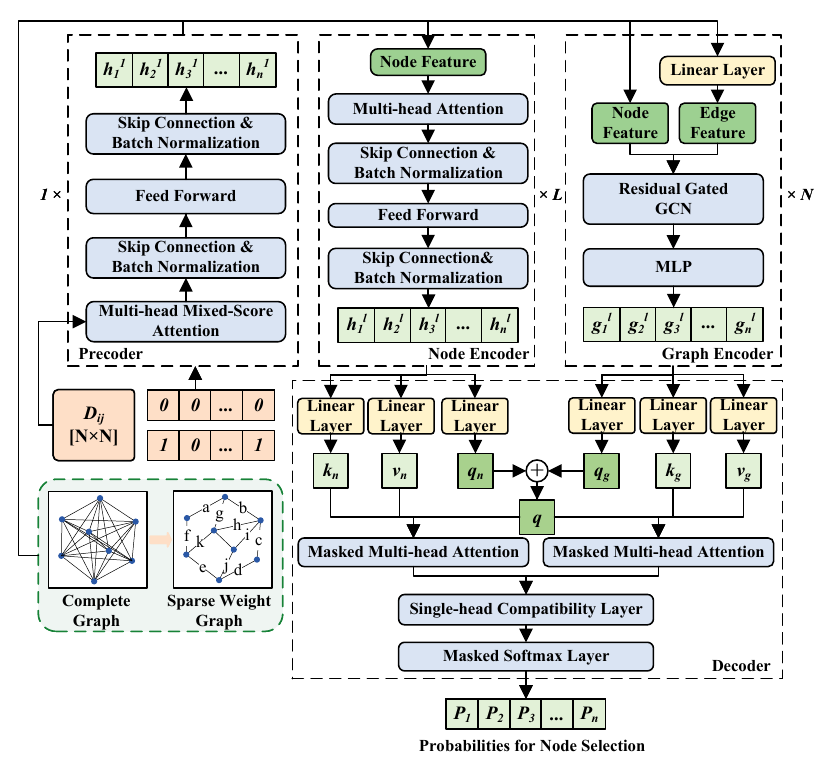}
    \caption{Detailed framework of our EFormer. From top to bottom: \emph{precoder}, \emph{node encoder}, \emph{graph encoder}, and \emph{decoder}. Select edges as the sole input to EFormer.}
    \label{fig:EFormer_detail}
\end{figure*}

Figure~\ref{fig:EFormer_detail} illustrates the detailed architecture of each module within our EFormer model. 
EFormer primarily consists of four essential components: the precoder, node encoder, graph encoder, and decoder. Specifically, the precoder is composed of a multi-head mixed-score attention layer and a feedforward (FF) layer, which is utilized to transform edge information into temporary node embeddings. The node encoder comprises six layers of multi-head attention (MHA) and FF, tasked with encoding the temporary node embeddings. The graph encoder consists of six graph convolutional network (GCN) layers and three multi-layer perceptron (MLP) layers, designed to process sparse graph embeddings. Through a parallel encoding strategy, the node encoder and graph encoder handle graph and node embeddings in distinct feature spaces, thereby providing a more comprehensive representation of the global relationships among edges.
Subsequently, during the decoding stage, two sets of keys, values, and queries are extracted from the two parallel encoded information. The attention mechanisms on the two encoded embeddings are computed via a dual-query integration approach, ultimately yielding the complete path construction.

\subsection{Graph Attention Networks (GATs)}
\label{appendix A.2}

The EFormer operates on a bipartite graph with weighted edges (which can also be a sparse graph). The graph consists of two identical sets of nodes, $N_x$ = $\{x_1,...,x_n\}$ and $N_y$= $\{y_1,...,y_n\}$. The edge between $x_i$ and $y_j$ has the weight $e(x_i, y_j) \equiv D_{ij}$, that is, the $(i, j)$-th element of matrix $D$. Here we show a symmetric matrix, but the EFormer is equally applicable to asymmetric matrices.
\begin{align}
\mathbf{D}_{ij} = \left( \begin{array}{cccc}
d_{11} & d_{12} & \cdots & d_{1n} \\
d_{21} & d_{22} & \cdots & d_{2n} \\
\vdots & \vdots & \ddots & \vdots \\
d_{n1} & d_{n2} & \cdots & d_{nn}
\end{array} \right)
\end{align} 

Traditional neural heuristics typically deal with fully connected nodes without edge weights, such as AM~\cite{kool2019AM} and POMO~\cite{kwon2020pomo}. Their encoders generally follow the graph attention networks (GATs) framework~\cite{velivckovic2017graph}. In each attention layer, the vector representation $\hat{h}_{x_i}$ of node $x_i$ is updated to $\hat{h}_{x_i}'$ using the aggregated representations of its neighbors as:
\begin{align}
\hat{h}_{x_i}' = \mathcal{F}\{\hat{h}_{x_i},{\hat{h}_{y_j}| y_j\in N_{x_i}}),
\end{align} 
where $N_{x_i}$ is the set of all neighbors of $x_i$. The learnable update function $\mathcal{F}$ consists of multi-head attention, whose aggregation process utilizes the attention scores of each pair of nodes $(x_i, y_j)$, which is a function of $\hat{h}_{x_i}$ and $\hat{h}_{y_j}$.

Our EFormer also generally follows the GATs framework. However, GATs computes attention scores for each node based solely on node features without incorporating edge weights into the attention mechanism. Our primary distinction from GATs lies in the explicit integration of edge weights into the update function. Specifically, the attention score of a pair of nodes $(x_i, y_j)$ is not only a function of $\hat{h}_{x_i}$ and $\hat{h}_{y_j}$, but also a function of the edge weight $e(x_i,y_j)$. EFormer is inspired by the encoder architecture of MatNet~\cite{kwon2021matrix}. However, unlike MatNet, which employs two distinct update functions, we only consider one update function per layer and focus on extracting effective node embeddings from the edge weights. The update function of EFormer can be described as: 
\begin{align}
\hat{h}_{x_i}' = \mathcal{F_R}\{\hat{h}_{x_i},\hat{h}_{y_j}, {e(x_i,y_j)|x_i \in R,y_j\in M}),
\end{align} 
where the learnable update function $\mathcal{F}$ consists of MHA and its aggregation process utilizes the attention scores of each pair of nodes $(x_i, y_j)$, which is a function of $\hat{h}_{x_i}$ and $\hat{h}_{y_j}$; $R$ denotes the set of adjacent nodes of $x_i$; and $M$ represents the set of adjacent nodes of $y_j$.


\begin{table*}[!t]
    \centering
    \resizebox{1.0\linewidth}{!}{
    \begin{tabular}{lrrrrrrrrrrrr}
        \toprule
          & \multicolumn{3}{c}{TSP100} & \multicolumn{3}{c}{TSP200}& \multicolumn{3}{c}{TSP300}& \multicolumn{3}{c}{TSP500}\\   
Method  & Len. & Gap(\%) & Time(m) & Len. & Gap(\%) & Time(m) & Len. & Gap(\%) & Time(m)& Len. & Gap & Time(m) \\\midrule    
Concorde& 7.731& 0.000 &8.42 &10.704& 0.000  &16.01 &12.934&0.000 &23.02 &16.522&0.000 &35.12
         \\  
        GCN-Greedy&8.457&9.919&0.07&12.324&15.141&0.28 &16.270&25.786&0.85&26.141&58.226&2.87 \\
        GCN-BS &8.111&7.506&0.08 &12.133&13.351&0.29 &15.964&23.424&0.85&25.686&55.468&2.87  \\
        GCN-BS*&7.903&2.232&0.64 &11.980&11.927&2.29 &15.841&22.474&5.33&25.364&53.518&23.69  \\
        
        MatNet(×1)&7.786&0.711&0.04 &11.395&6.459&0.16 &14.930&15.431&0.72 &22.511&36.253&2.12  \\
        MatNet(×8)&7.747&0.213&0.08 &11.282&5.408&0.35 &14.766&14.159&1.08 &22.199&34.364&3.89  \\
        \midrule    

       EFormer(×1)&7.754&0.302&0.04 &11.328&5.837&0.25 &15.077&16.563&1.02 &21.281&28.807&2.12  \\
        EFormer(×8)&7.739&\textbf{0.113}&0.11 &11.136&\textbf{4.043}&0.42 &14.508&\textbf{12.170}&1.86 &20.397&\textbf{23.457}&5.68 \\
    \midrule \midrule
     & \multicolumn{3}{c}{CVRP100} & \multicolumn{3}{c}{CVRP200}& \multicolumn{3}{c}{CVRP300}& \multicolumn{3}{c}{CVRP500}\\       
 Method & Len. & Gap(\%) & Time(m) & Len. & Gap(\%) & Time(m) & Len. & Gap(\%) & Time(m) & Len. & Gap(\%) & Time(m)   \\
        \midrule    
        
LKH3  &15.594 & 0.000 &1.11h &20.173& 0.000&2.5h &26.028 &0.000&3.6h&37.229 &0.000 &5.8h \\  

        GCN-Greedy&16.860&8.112&0.10&22.606&12.063&0.28 &33.889&30.202&1.33&49.550&33.096&5.64 \\
        GCN-BS &16.442&5.438&0.07&22.282&10.454&0.28 &33.213&27.605&1.87&48.923&31.410&3.51\\
        GCN-BS*&16.128&3.427&1.45&22.066&9.387&1.85 &32.802&26.025&4.17&48.311&29.767&9.01 \\
        
        MatNet(×1)&16.220&4.017&0.05&21.982&8.971&0.14&31.069&19.365&0.34&46.908&25.999&1.95 \\
        MatNet(×8)&16.075&3.086&0.09&21.715&7.648&0.34&30.448&16.982&0.94&46.107&23.845&3.59 \\
        \midrule    
      
        EFormer(×1)&15.809&1.381&0.06 &20.831&3.266&0.16 &28.211&8.387 &0.43 &42.708 &14.716&2.13 \\
        EFormer(×8)&15.730&\textbf{0.876}&0.10 &20.665&\textbf{2.443}&0.47 &27.723&\textbf{6.511} &1.27 &41.647 &\textbf{11.865}&5.44\\
        \bottomrule
    \end{tabular}
    }
    \caption{Experiment results of large-scale generalization on TSP and CVRP.}
    \label{tab3:vrp-scale-general}
\end{table*}

\begin{table*}[!t]
    \centering
    \resizebox{0.81\linewidth}{!}{
    \begin{tabular}{lrrrrrrrrrr}
        \toprule
          & \multicolumn{3}{c}{TSP20} & \multicolumn{3}{c}{TSP50}& \multicolumn{3}{c}{TSP100}\\         
        Method  & Len. & Gap(\%) & Time(m) & Len. & Gap(\%) & Time(m) & Len. & Gap(\%) & Time(m)    \\\hline 
         \multirow{1}*{Distribution} & \multicolumn{9}{c}{Explosion}   \\
         \midrule 
        Concorde&3.559& 0.000&3.43 &4.620 & 0.000 &21.53 &5.830 &0.000&66.11 \\
        GCN Greedy&3.673&3.180&0.46 &4.931 &6.610 &1.24 &6.491 &11.340&7.88 \\
        GCN-BS &3.591&0.898&0.44 &4.686 &1.43 &1.21 &6.388 &9.564 &8.48  \\
        GCN-BS*&3.560&0.008&34.85 &4.645&0.546&41.29 &6.247 &7.144 &78.05 \\
        MatNet(×1)&3.561&0.044&0.11 &4.643&0.495&0.13&5.981&2.587&0.52 \\
        MatNet(×8)&3.560&0.001&0.33 &4.626&0.121&1.91 &5.915 &1.450&5.28  \\
    EFormer(×1)&3.560 &0.022 &0.04 &4.630 &0.225&0.26 &5.905&1.293& 1.26 \\  
    EFormer(×8)&3.559&\textbf{0.000} &0.24 &4.623 &\textbf{0.058}&1.34 &5.864&\textbf{0.586}& 6.21     \\
     \midrule 
     
        \multirow{1}*{Distribution} & \multicolumn{9}{c}{Grid}   \\
         \midrule 
        Concorde&3.845& 0.000 &3.43& 5.969 & 0.000&21.43 &7.790 &0.000 &66.23 \\
        GCN Greedy&3.957&2.912&0.46&5.970&4.915&1.02 &8.708&11.781 &4.09\\
        GCN-BS &3.876&0.816&0.44&5.732 &0.745&1.18 &8.322&6.826 &4.31 \\
        GCN-BS*&3.845&0.004&34.29&5.700&0.167 &34.14&8.171&4.892 &61.15 \\
        MatNet(×1)&3.847&0.041&0.11 &5.709&0.336 &0.13 &7.917&1.633&0.52 \\
        MatNet(×8)&3.845&0.001&0.34 &5.695&0.085&1.24 &7.853&0.805&6.12\\
        EFormer(×1)&3.846 &0.017&0.04 &5.700 &0.167&0.26 &7.815&0.325&1.26\\
    EFormer(×8)&3.845&\textbf{0.000}&0.24 &5.693 &\textbf{0.055}&1.34 &7.799&\textbf{0.110}&6.17\\ 
          \midrule 
        \multirow{1}*{Distribution} & \multicolumn{9}{c}{Implosion}   \\
         \midrule 
        Concorde&3.785& 0.000&2.48 &5.600& 0.000&21.64 &7.610&0.000&66.45 \\
        GCN Greedy&3.899&3.024&0.47&5.881&5.02&1.15 &8.636 &13.476&4.75 \\
        GCN-BS &3.814&0.771&0.43&5.645 &0.797&1.21 &8.285 &8.866 &5.59 \\
        GCN-BS*&3.785&0.004&34.61&5.610&0.176&35.46 &8.137&6.919&60.45 \\
        MatNet(×1)&3.786&0.457&0.11&5.620&0.351&0.13 &7.740&1.709&0.52 \\
        MatNet(×8)&3.785&0.006&0.34&5.605&0.092&1.91 &7.675&0.850&4.03\\
        
       EFormer(×1)&3.785&0.019&0.04&5.609&0.167&0.26 &7.636&0.335&1.26\\
    EFormer(×8)&3.785&\textbf{0.001}&0.23 &5.603 &\textbf{0.056}&1.31 &7.619&\textbf{0.122}&6.32 \\ 
        \midrule \midrule
        
       & \multicolumn{3}{c}{CVRP20} & \multicolumn{3}{c}{CVRP50}& \multicolumn{3}{c}{CVRP100}\\  
        Method  & Len. & Gap(\%) & Time(m) & Len. & Gap(\%) & Time(m) & Len. & Gap(\%) & Time(m)    \\\hline 
        \multirow{1}*{Distribution} & \multicolumn{9}{c}{Explosion}   \\
         \midrule 
        LKH3  &5.748 &0.000&2.35h &8.740& 0.000 &8.52h &12.302&0.000&13.46h    \\
        GCN Greedy&5.874&2.191&0.24 &9.414&7.714&1.64 &13.394&8.713&4.57 \\
        GCN-BS &5.836&1.537&0.24 &9.258&5.927&1.95&13.176&6.949&5.78 \\
        GCN-BS*&5.795&0.826&20.85 &8.972&2.651&45.46 &12.988 &5.425 &65.51  \\
MatNet(×1)&5.806&1.020&0.11&9.131&4.470&0.23&12.933&4.979&0.88 
        \\
        MatNet(×8)&5.780&0.568&0.52 &8.985&2.803&1.13&12.762&3.585&4.70   \\
       EFormer(×1)&5.783&0.616 &0.05 &8.834 &1.078&0.25 &12.535 &1.749& 0.98 \\
        EFormer(×8)&5.758&\textbf{0.183} &0.28 &8.792 &\textbf{0.595}&1.69 &12.458 &\textbf{1.123}& 6.86 \\\hline 
        \multirow{1}*{Distribution} & \multicolumn{9}{c}{Grid}   \\
         \midrule 
        LKH3    &6.147 &0.000&2.15h &10.400& 0.000 &8.52h &15.640 &0.000 &13.46h \\ 
         
        GCN Greedy&6.432&4.633&0.49&10.915 &4.953&1.73 &16.432 &5.065 &3.93 \\
        GCN-BS &6.370 &3.627&0.38 &10.723 &3.104 &1.82 &16.257 &3.947 &4.18   \\
        GCN-BS*&6.313 &2.700&21.18&10.657 &2.466 &44.49&16.121 &3.074 &73.07 \\
        
        MatNet(×1)&6.208&0.978&0.15 &10.832&4.153 &0.34 &16.282&4.104&0.52   \\
        MatNet(×8)&6.181&0.538&0.71 &10.678&2.677 &1.68 &16.115&3.036&4.65     \\
        
        EFormer(×1)&6.185&0.605 &0.05 &10.502 &0.984&0.25 &15.842 &1.291 & 0.98     \\
        EFormer(×8)&6.159&\textbf{0.180}& 0.30 &10.459 &\textbf{0.569}&1.65 &15.773&\textbf{0.853}&6.86      \\\hline 

        \multirow{1}*{Distribution} & \multicolumn{9}{c}{Implosion}   \\
         \midrule 
        LKH3    &6.061 & 0.000 &2.15h &10.260 & 0.000 &8.52h &15.440 &0.000 &13.46h \\
         
        GCN Greedy&6.247&3.072&0.24&10.755 &4.824&1.47 &16.297&5.549 & 4.46 \\
        GCN-BS &6.184 &2.037&0.24  &10.676 &4.056&1.52 &16.023&3.776& 4.63  \\
        GCN-BS*&6.133 &1.184&20.84&10.547 &2.793&45.09&15.928&3.162 & 65.87 \\
        
        MatNet(×1)&6.121&0.992 &0.09 &10.691&4.196&0.13 &16.080&4.148&0.78   \\
        MatNet(×8)&6.095&0.557&0.51 &10.540&2.732 &1.18 &15.917&3.089&4.74  \\
        EFormer(×1)&6.098&0.610&0.06 &10.362&0.993&0.25&15.644 &1.323 & 0.98   \\
        EFormer(×8)&6.071&\textbf{0.176}&0.32 &10.319&\textbf{0.575}&1.72&15.574 &\textbf{0.871} & 6.86 \\  
        \bottomrule
    \end{tabular}
    }
    \caption{Experiment results of three distribution generalizations on TSP and CVRP. The three distributions are explosion, grid and implosion.}
    \label{tab4:vrp-distrib}
\end{table*}

\section{EFormer Related Experiments}
\label{appendix B}

We introduce the specific experimental settings and related experimental results, including datasets, baselines, hyperparameter, and generalization experiments.
 
\subsection{Experiment Setting}
\label{appendix B.1}
 
\paragraph{Problem Setting.}
We follow the standard data generation procedures from prior work~\cite{kool2019AM} to create training and testing datasets for TSP and CVRP, where distances between nodes are calculated and provided as inputs accordingly. Each training epoch samples 100,000 random instances, while a separate set of 10,000 uniformly generated instances is used for testing. Optimal solutions for TSP are obtained via the Concorde solver~\cite{cook2011traveling}, and those for CVRP using LKH3~\cite{helsgaun2017extension}. 

\paragraph{Baselines.}
We compare our EFormer with: 

\textbf{(1) Classical solvers:} Concorde~\cite{cook2011traveling}, LKH3~\cite{helsgaun2017extension}, HGS~\cite{vidal2022hybrid}, and OR-Tools~\cite{ORTOOLS}; 

\textbf{(2) Heatmap-based method:} GCN-BS~\cite{joshi2019efficient}; 

\textbf{(3) Edge-based neural heuristics:} MatNet~\cite{kwon2021matrix} and GREAT~\cite{lischka2024great}. 

The Baselines description is as follows:

1) GCN-BS takes both node and edge information as model input in the original paper. We modify it to edge-only input and keep other parts the same as the original model setting, where node features are obtained from edges. Finally, we train it with the same model structure and parameters so that the edge-based GCN-BS can be used to solve TSP and CVRP.

2) MatNet is used to solve ATSP and FFSP in their original paper, but in the appendix, they also solved the symmetric TSP with Euclidean distance. Therefore, we follow this method to retrain and test MatNet on symmetric TSP instances. In addition, we have improved and extended MatNet to solve CVRP, enhancing its applicability to a broader range of routing problems.

3) GREAT: Since the code is not publicly available, we directly use the results in their paper for comparison.

\paragraph{Model Setting.}
The EFormer model is a heavy encoder light decoder structure. We use an identical set of model hyperparameters across all problem sizes.
Precoder consists of an attention layer and a feed-forward layer. Graph encoder consists of $l_{conv} = 6$ GCN layers and $l_{mlp} = 3$ MLP layers with hidden dimension $h = 256$ for each layer. We consider $k = 20$ nearest neighbors for each node in the adjacency matrix $W_{adj}$. The node encoder consists of $l_{attention} = 6$ attention layers. 
In each attention layer, the head number of MHA is set to 16, the embedding dimension is set to 256, and the feedforward layer dimension is set to 512.
In all experiments, we train and test our models using a single Tesla V100-SXM2-32GB GPU.

\paragraph{Training.}
Both TSP and CVRP models are trained on randomly sampled instances, respectively. During each training epoch, 100,000 instances are randomly sampled. The optimizer is Adam~\cite{kingma2014adam} with an initial learning rate of 1e-4. The value of the learning rate decay is set to 0.1 for the TSP and CVRP. The TSP20 and TSP50 models are trained for 510 epochs, while the CVRP model is trained for 1010 epochs. For TSP100, the training duration is also set to 1010 epochs to achieve better results.
We employ the same training algorithm as POMO to train our model through reinforcement learning. Similarly, for different scales of instance models, $N$ different solutions are generated starting from each node. The average distance length of these $N$ solutions is used as the baseline of the algorithm~\cite{williams1992simple}.

\paragraph{Metrics and Inference.}
We report the optimality gap and inference time for all methods. The optimality gap measures the difference between the solutions obtained by various learning-based and non-learning methods and the optimal (or best) solutions obtained using Concorde for TSP and LKH3 for CVRP.
For GCN-BS and MatNet, as their pre-trained models are not available, we retrain all those models (including N = 20, 50, and 100) ourselves and report the comparative results on our test set. 
We use the POMO inference algorithm~\cite{kwon2020pomo}. That is, we allow the decoder to generate $N$ solutions in parallel, each starting from a different node, and choose the best solution. For our EFormer model, we report the greedy (×1) inference results as well as the instance augmentation results (×8 and ×128). 
 

\subsection{Generalization to larger-scale instances}
\label{appendix B.2}
Table~\ref{tab3:vrp-scale-general} shows the performance of EFormer on TSP and CVRP instances with up to 500 nodes. We evaluate the generalization performance of the GCN-BS, MatNet, and EFormer models with different inference strategies using a uniform test set of 128 instances. Despite the device limitations, the experimental results indicate that EFormer outperforms other edge-based methods on larger-scale instances, highlighting its strong generalization ability even when only edge information is available.

\subsection{Generalization across different distributions}
\label{appendix B.3}
We have reported the performance of the EFormer on uniformly distributed instances in the main text. Table~\ref{tab4:vrp-distrib} further evaluates the performance of EFormer on TSP and CVRP with explosion, grid, and implosion distributions. For the test dataset, we utilize 10,000 instances with three distributions for generalization testing. The results indicate that the EFormer model consistently outperforms GCN-BS and MatNet on all three distributions. Specifically, EFormer achieves the best performance on grid-distributed instances and ranks second on implosion-distributed instances. Moreover, EFormer maintains robust performance not only on the uniform distribution but also under varying distribution scenarios, thereby underscoring its exceptional generalization capabilities.

\begin{figure*}[!t]
    \centering
    \includegraphics[width=1.0 \linewidth]{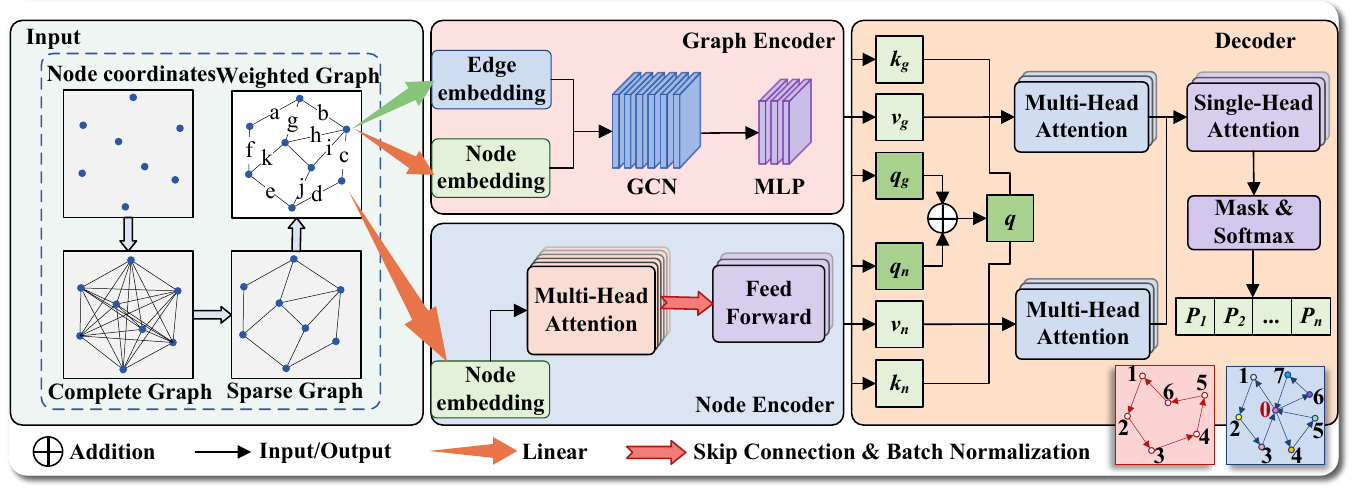}
    \caption{Overall architecture of our EFormer-node scheme.
    From the input of sparse graph (including nodes and edges), the \emph{graph encoder} and \emph{node encoder} operate in parallel, and finally the \emph{decoder} constructs the complete path.}
    \label{fig:EFormer-node-overall}
\end{figure*}

\begin{table*}[!t]
    \centering
    \resizebox{1.0\linewidth}{!}{
    \begin{tabular}{lrrrrrrrrr}
        \toprule
        & \multicolumn{3}{c}{TSP20} & \multicolumn{3}{c}{TSP50}& \multicolumn{3}{c}{TSP100}\\  
        Method  & Len. & Gap(\%) & Time(m) & Len. & Gap(\%) & Time(m) & Len. & Gap(\%) & Time(m)    \\
        \midrule    
        Concorde&3.831&0.000&4.43 &5.692&0.000 &23.53&7.763 &0.000&66.45\\
        LKH3  &3.831&0.000&2.78 &5.692&0.000&17.21&7.763 &0.001&49.56\\
        OR-Tools&3.864&0.864&1.16&5.851& 2.795&10.75&8.057&3.782&39.05 \\\hline
        
        GCN Greedy&3.856&0.650&0.25 &5.901&3.678&1.21&8.413&8.373&6.25\\
        GCN-BS   &3.835&0.128&0.81 &5.710&0.317&4.62 &7.931&2.155&17.73\\
        GCN-BS*&3.831 &0.000&21.24 &5.694&0.041&37.63&7.869&1.368&58.34\\
        POMO(×1) &3.832 &0.026&0.08&5.698&0.119&0.24 &7.796&0.364&1.03 \\
        POMO(×8)&3.831 &0.001 &0.12&5.693&0.024&0.45 &7.775&0.142&2.11 \\

        Pointerformer(×1) &3.833&0.067&0.09&5.714&0.391&0.19 &7.859&1.229&0.67\\
        Pointerformer(×8)&3.831&0.001 &0.18&5.693&0.026&0.52&7.776&0.163&1.51\\
        LEHD Greedy&3.867&0.961&0.14&5.721&0.519&0.24 &7.808&0.577 &1.02\\
        LEHD RRC10 &3.834&0.091&0.35&5.697&0.095&1.10&7.775 &0.145 &2.12 \\
        LEHD RRC50 &3.831&0.004&0.62&5.692&0.010&2.45&7.766 &0.033 &7.93 \\\hline

       EFormer(×1)&3.831&0.018&0.04 &5.699 &0.130&0.26 &7.788&0.324& 1.22\\
       EFormer(×8)&3.831&0.000&0.15 &5.692 &0.011&1.34 &7.772&0.115& 6.81\\
        EFormer(×128) &3.831 &\textbf{0.000}&4.72 &5.692&\textbf{0.000}&25.81&7.767 &\textbf{0.045}&66.55 \\\hline

        EFormer-node(×1)&3.831 &0.008&0.04&5.695 &0.055&0.24&7.772&0.113&1.20\\
        EFormer-node(×8)&3.831 &\textbf{0.000} &0.18&5.692 &\textbf{0.005}&1.59&7.765&\textbf{0.027}&6.73\\

        \midrule   \midrule 
        & \multicolumn{3}{c}{CVRP20} & \multicolumn{3}{c}{CVRP50}& \multicolumn{3}{c}{CVRP100}\\  
        Method  & Len. & Gap(\%) & Time(m) & Len. & Gap(\%) & Time(m) & Len. & Gap(\%) & Time(m)    \\
        \midrule  
        LKH3&6.117&0.000&2.15h&10.347&0.000&8.52h &15.647 &0.000 &13.46h\\
        HGS&6.112&-0.079&1.48h&10.347&-0.001 &4.67h &15.584&-0.401&6.54h\\
    OR-Tools&6.414&4.863&2.37&11.219&8.430&19.35&17.172&9.749&2.61h \\\hline
        POMO(×1)&6.160&0.698&0.06&10.533&1.799&0.18&15.837&1.216&0.64 \\
        POMO(×8)&6.132&0.254&0.20&10.437&0.875&0.48&15.754&0.689&2.11 \\

        LEHD Greedy&6.462&5.647&0.07&10.872&5.075&0.18&16.217&3.648&0.55 \\
        LEHD RRC10 &6.222&1.718&0.31&10.586&2.307&0.69&15.885 &1.529 &2.33  \\
        LEHD RRC50 &6.140&0.375&1.57&10.430&0.803&2.95&15.729 &0.524&9.10
        \\\hline

EFormer(×1)&6.147&0.490&0.04&10.457&1.067&0.25&15.844&1.259&0.98 \\
EFormer(×8)&6.123&0.098&0.28&10.414&0.650&1.69&15.776&0.830 &6.86\\
EFormer(×128)&6.116&\textbf{-0.017}&14.84&10.393&\textbf{0.447}&24.47&15.735&\textbf{0.563} &85.65  \\ \hline
    
  EFormer-node(×1)&6.141&0.388&0.06&10.455&1.045&0.22&15.795&0.950&0.96\\
  EFormer-node(×8)&6.124&\textbf{0.114}&0.27&10.411&\textbf{0.625}&1.54
            &15.725&\textbf{0.498}&6.84\\

        \bottomrule
    \end{tabular}
    }
    \caption{Experimental results of our EFormer-node on TSP and CVRP with uniformly distributed instances.}
    \label{tab:vrp-node}
\end{table*}


\section{Node-based EFormer}
\label{appendix C}

\subsection{EFormer-node Method}
\label{appendix C.1}
To facilitate a comprehensive comparison between EFormer-based solvers and other neural heuristic solvers, we evaluate our model using traditional node coordinates. We introduce a variant of the EFormer model, termed EFormer-node, which takes node coordinates as input. This variant retains the core structural framework of the edge-based EFormer model described in the main text but incorporates a key modification: the omission of the precoder module. The precoder is conventionally used to convert edge information into node embeddings. By excluding this module, EFormer-node directly utilizes the original node coordinates as input, streamlining the model architecture and enhancing computational efficiency. The specific experimental setup for EFormer-node is consistent with that of the original EFormer model, with detailed descriptions provided in Appendix~\ref{appendix B.1}.

The comprehensive architecture of the EFormer-node model is illustrated in Figure~\ref{fig:EFormer-node-overall}. In this model, node coordinates and edge weight matrix are directly utilized as inputs to test the EFormer-node, thereby eliminating the need for the precoder module. This streamlined approach enhances the model's efficiency and adaptability.
The EFormer-node structure is composed of three integral components: graph encoder, node encoder, and decoder. These components are consistent with the EFormer model described in the main text, ensuring consistency and compatibility. Specifically, the graph encoder is equipped with 6 GCN layers, while the node encoder comprises 6 attention layers.
By incorporating node coordinates and a sparse graph, EFormer-node adeptly encodes these elements into distinct feature spaces. This dual encoding process results in two global relationship representations that integrate node and edge information. This integration is pivotal for capturing the intricate relationships within the instances, thereby enhancing the model's representational capacity. In the decoder phase, parallel context embedding and multi-query integration are used to compute separate attention mechanisms over the two encoded embeddings, facilitating efficient path construction. 

The provision of actual node coordinates equips the model with a more comprehensive set of global encoding information. This enriched information base significantly outperforms models that rely solely on edge information, thereby elevating the EFormer-node model to a higher standard of performance. By incorporating the actual node coordinates, the model acquires a more comprehensive set of global encoding information, thereby achieving superior performance compared to models that rely solely on edge information.

\subsection{EFormer-node Experimental Setting}
\label{appendix C.2}

\paragraph{Model Setting.}
We adopt the same experimental settings as the EFormer model described in the main text, encompassing datasets, hyperparameters, training algorithms, and inference techniques. For detailed experimental settings, please refer to Appendix~\ref{appendix B.1}.
The TSP and CVRP datasets are randomly generated with nodes uniformly distributed, and the problem sizes are set to $N$ = 20, 50, and 100. In each training phase, 100,000 instances are randomly generated. We utilize the same test set as the EFormer model in the main text for validation.

For the EFormer-node model, we remove the precoder part while retaining the graph encoder and node encoder. The corresponding model structure and parameter settings are consistent with the experiments mentioned in the main text.
For TSP20 and TSP50, we train the models for 510 epochs, whereas for TSP100 and all CVRP experiments, we train for 1010 epochs. We employ the same training algorithm as POMO to train our model through reinforcement learning.

\paragraph{Baselines.}
We compare our EFormer-node with 

\textbf{(1) Classical solvers:} Concorde~\cite{cook2011traveling}, LKH3~\cite{helsgaun2017extension}, HGS~\cite{vidal2022hybrid} and OR-Tools~\cite{ORTOOLS};

\textbf{(2) Heatmap-based method:} GCN-BS~\cite{joshi2019efficient}; 

\textbf{(3) Node-based neural heuristics:} POMO~\cite{kwon2020pomo}, Pointerformer~\cite{jin2023pointerformer}, LEHD~\cite{luo2023neural}.

\subsection{EFormer-node Experimental Results}
Table~\ref{tab:vrp-node} shows the main experimental results of our EFormer-node model on uniformly distributed TSP and CVRP instances.
The first group of baselines for comparison are the results of several representative solvers, including Concorde, LKH3, HGS, and OR-Tools. Consistent with the experimental results reported in the main text, we use the Concorde solver to obtain the optimal solution for the TSP test set and the LKH3 solver to obtain the optimal (or best) solution for the CVRP test set.
The second group of baselines included in the table comprise popular node-based neural heuristics, such as POMO and LEHD. For TSP, we also compare our results with those obtained using the GCN and Pointerformer methods. To demonstrate the effectiveness of our EFormer framework, we report not only the performance of the node-based EFormer model (i.e., EFormer-node) but also the results of the edge-based EFormer model (i.e., EFormer), which aligns with the results presented in the main text. For all methods, we report the optimality gap and inference time.

For TSP, the table clearly demonstrates that our EFormer-node method markedly surpasses all other methods in both greedy and ×8 instance augmentation inference performance. Notably, within the same inference time, we compare our EFormer-node with LEHD using 50 RRC iterations and find that our EFormer-node consistently performs better. This superiority is further highlighted on TSP100, where our model achieves an impressively low optimality gap of merely 0.0269\%. 
In addition, it is evident that, with the exception of LEHD RRC50, our edge-based EFormer model (as described in the main text) outperforms other node-based neural heuristics, including POMO, Pointerformer, and even LEHD RRC10, in terms of both greedy and ×8 instance augmentation inference performance. These results fully demonstrate the highly competitive performance of our proposed EFormer framework across diverse inputs.

For CVRP, our EFormer-node also achieves promising performance in both greedy and ×8 instance augmentation inference for all instances. Among all the neural heuristic methods compared, our EFormer-node model emerges as the frontrunner, achieving optimal performance in both greedy and ×8 instance augmentation inference. Notably, even when compared with LEHD RRC50, our method also demonstrates significant advantages. For CVRP100, the average optimality gap of our EFormer-node model is impressively reduced to 0.4731\%, highlighting its exceptional performance in solving CVRP.
Furthermore, for the edge-based EFormer presented in the main text, we observe that it performs slightly worse than LEHD RRC50 on CVRP100. However, besides this, our edge-based EFormer outperforms other node-based neural methods on other problem instances.

Evidently, the EFormer-node exhibits superior greedy inference performance and significantly outperforms POMO and LEHD across all instances when augmented with ×8 instance enhancement. Concurrently, the edge-based EFormer also demonstrates remarkable performance, even surpassing certain node-based methods. Collectively, these experimental results underscore the robustness of our proposed EFormer framework, which delivers strong performance in both edge-based and node-based contexts. These findings strongly demonstrate the exceptional performance of our proposed EFormer framework across diverse inputs, thereby highlighting its extensive adaptability and versatility.


\section{Asymmetric Traveling Salesman Problem (ATSP)}
\label{appendix D}
 
\begin{table}[!t]
    \centering
    \resizebox{0.90\linewidth}{!}{
    \begin{tabular}{lrrr}
        \toprule
         & \multicolumn{3}{c}{ATSP50}  \\
          Method & Len. & Gap(\%)  & time(m)    \\
        \midrule    
        CPLEX &1.559& 0.000  & 60.05 \\ 
        
Nearest Neighbor& 2.100 &34.610 & -  \\ 
Nearest Insertion&1.950 &25.160 & - \\ 
Furthest Insertion&1.840 &18.220 &-  \\  
LKH3    &1.560 &	0.000 & 0.19 \\ 
        \midrule    
MatNet(×1) &1.580 &1.350&0.15  \\ 
MatNet(×8) &1.566 &0.472&1.23   \\ 
MatNet(×128)&1.561 &0.144&16.55 \\ 
        \midrule    
EFormer(×1)&1.576 &\textbf{1.105}  & 0.27 \\ 
EFormer(×8)&1.565&\textbf{0.399 }& 1.65  \\ 
EFormer(×128)&1.561&\textbf{0.143} &25.16   \\ 

        \bottomrule
    \end{tabular}
    }
    \caption{Experimental results on ATSP50 with 10,000 instances.}
    \label{tab:ATSP}
\end{table}

Our EFormer, which takes edge information as input, can be readily extended to address the Asymmetric Traveling Salesman Problem (ATSP). To evaluate the generalizability of the EFormer-based solver against other methods, we test its performance on ATSP instances. We utilize the same model architecture and parameter settings as described in the main text for the TSP experiments, with the sole difference being the training dataset. While the TSP experiments use symmetric distance matrices, for ATSP, we employ randomly generated asymmetric distance matrices as inputs. Additionally, both the training and testing datasets follow the data generation method of MatNet.

We present comparative experimental results for different methods on ATSP50 instances in Table~\ref{tab:ATSP}. One of the key baselines compared in the table is the MatNet model. Similar to our EFormer, MatNet is an edge-based construction method specifically designed for solving ATSP. 
The results in Table~\ref{tab:ATSP} demonstrate that our EFormer-based method achieves competitive performance on ATSP, showcasing its wide adaptability. This further substantiates the effectiveness of the EFormer architecture in addressing diverse problems.

\end{document}